\documentclass[
]{ceurart}

\usepackage{amsmath}
\usepackage{amssymb}
\usepackage{todonotes}
\usepackage{tikz}
\usepackage{subcaption}
\usepackage{csquotes}
\usepackage{xcolor}
\usepackage{framed}
\usepackage{moreverb}
\usepackage{cleveref}
\usepackage{enumitem}
\usetikzlibrary{calc, decorations.pathreplacing}
\usepackage{adjustbox}

\usepackage{listings}
\usepackage{graphbox}
\DeclareRobustCommand{\DLLogo}{%
  \begingroup\normalfont
  \kern-1.75pt\includegraphics[align=c,height=1.25\baselineskip]{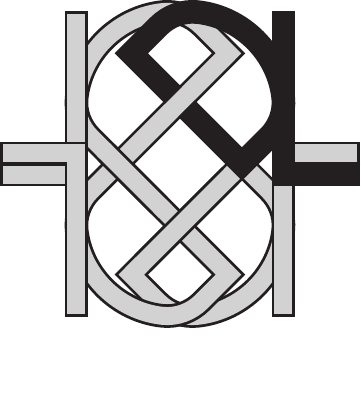}\kern-1.5pt%
  \endgroup
}

\usepackage{amsthm}

\newtheorem{definition}{Definition}
\newtheorem{example}{Example}

\newcommand{\patrick}[1]{\todo[inline,color=pink]{{\footnotesize Patrick: #1}}}

\DeclareMathOperator{\pre}{\textsf{pre}}
\DeclareMathOperator{\eff}{\textsf{eff}}
\DeclareMathOperator{\add}{\textsf{add}}
\DeclareMathOperator{\del}{\textsf{del}}

\newcommand{\OWLcolor}{MidnightBlue}
\newcommand{\Planningcolor}{BrickRed}
\newcommand{\Interfacecolor}{OliveGreen}
\tikzstyle{textNode} = [
	anchor=center,
	]

\tikzstyle{class} = [
	textNode,
	rounded corners,
	minimum height=16pt,
	minimum width=20pt,
	line width=0.75,
	draw
	]
\tikzstyle{individual} = [
	draw=black,
	fill opacity=0.1,
	text opacity=1,
	textNode,
	rounded corners,
	minimum height=16pt,
	minimum width=20pt,
	line width=0.75
	]

\tikzstyle{relation} = [
	line width=0.75pt,
	]

\newcommand{\Omc}{\ensuremath{\mathcal{O}}\xspace}
\newcommand{\tup}[1]{\langle #1 \rangle}
\newcommand{\PDDLcomponent}{\textbf{P}\xspace}
\newcommand{\PDDLdomain}{\ensuremath{D}\xspace}
\newcommand{\PDDLproblem}{\ensuremath{P}\xspace}
\newcommand{\Ind}{\textsf{Ind}}
\newcommand{\Dmc}{\mathcal{D}}

\newcommand{\HornALCHOIQ}{\text{Horn-}\ensuremath{\mathcal{ALCHOIQ}}\xspace}

\newcommand{\pddl}[1]{\textup{\texttt{#1}}}
\newcommand{\owl}[1]{\textup{\textsf{#1}}}

\newcommand{\HornImpl}{\textsf{\textsc{Horn}}\xspace}
\newcommand{\InterfaceImpl}{\textsf{\textsc{OM}}\xspace}

\begin{document}

\copyrightyear{2023}
\copyrightclause{Copyright for this paper by its authors.
	Use permitted under Creative Commons License Attribution 4.0
	International (CC BY 4.0).}

\conference{}

\title{Towards Ontology-Mediated Planning with OWL~DL~Ontologies (Extended Version)}

\author[1]{Tobias John}[
orcid=0000-0001-5855-6632
]
\address[1]{University of Oslo, Gaustadall\'{e}en 23B, 0316 Oslo, Norway}
\author[2]{Patrick Koopmann}[%
orcid=0000-0001-5999-2583
]
\address[2]{Vrije Universiteit Amsterdam, De Boelelaan 1105, 1081 HV Amsterdam,
	The Netherlands}

\begin{abstract}
While classical planning languages make the closed-domain and closed-world assumption, there have been various
approaches to extend those with DL reasoning, which is then interpreted under the usual open-world semantics.
Current approaches for planning with DL ontologies integrate the DL directly into the planning language, and practical
approaches have been developed based on first-order rewritings or rewritings into datalog. We present here a new
approach in which the planning specification and ontology are kept separate, and are linked together using an interface.
This allows planning experts to work in a familiar formalism, while existing ontologies can be easily integrated and
extended by ontology experts. Our approach for planning with those ontology-mediated planning problems
is optimized for cases with comparatively small domains, and supports the whole OWL DL fragment. The idea is to
rewrite the ontology-mediated planning problem into a classical planning problem to be processed by existing planning
tools. Different to other approaches, our rewriting is data-dependent. A first experimental evaluation of our
approach shows the potential and limitations of this approach.
\end{abstract}

\begin{keywords}
Planning \sep
OWL Ontologies \sep
Description Logics
\end{keywords}

\maketitle

\section{Introduction}

We present a new formalism to integrate OWL ontologies into planning problems, together with a first
practical technique for automated planning for such ontology-mediated planning problems. Different
to existing approaches, our formalism keeps the ontology component and the planning component separate
from each other. Our practical implementation is optimized for planning problems with small domains,
and is a first technique for automated planning that supports full OWL.

Both planning and ontologies are commonly
used in approaches to develop autonomous
robots~\cite{OlivaresAlarcosBKGH2019,KarpasM2020}.
The motivation for the present work comes from planning problems for
autonomous underwater vehicles
(AUVs).
Such robots are often used for inspection tasks, e.g. of underwater
infrastructure such as pipelines or oil platforms, as well as for mapping of the sea floor
\cite{SahooDR2019}, but eventually they should also be able to complete
%
more
complex missions that include manipulation tasks \cite{CashmoreFLMR2015}. The
robots need to be able to work autonomously, because their operation area is very
remote and without a connection to a human operator. Even recovering the
vehicle
in case of a problem is a difficult and time consuming task. Therefore, the
mission plans for such vehicles should be as robust as possible, which includes
that the robots have some understanding of the domain they operate in.
This domain knowledge is not specific to planning, and would thus be ideally formalized 
in an ontology that can also be used in other contexts of AUVs, such as 
configuring them,
or recognizing unexpected situations~\cite{SITUATION_RECOGNITION_OWL}.
For example, such an ontology might define a concept of ProtectedAnimal, based
on
the concept of Animal and having a position that is located in a 
NatureProtectionArea. Using
such an ontology, the robot would then be able to understand when it needs to 
keep a larger distance to an animal in order not to disturb it.
Ontologies are an ideal framework to represent such domain knowledge, and there 
are existing ontologies for the underwater domain, 
such as
the SWARMS
ontology \cite{LiBMBR2017,ZhaiMLC2018}. 
However, if we want to use such an ontology in connection with planning, we need 
a planning framework that can make use of the ontology.
%


In this paper, we propose
a general framework to connect planning problems with
OWL ontologies, and a technique to compute plans for such problems. Using this
framework, we can create a planning domain that interacts
with the ontology to generate plans that take its domain knowledge into
account. Similar to \cite{CashmoreFLMR2015}, we use the ontology to model the
environment. But additionally, we model actions of the robot that
manipulate the objects and the relations between objects in the environment, 
e.g. that the
robot opens or closes a valve.

\label{sec:related-work}

Using ontologies to support planning is not a new idea, and has been
investigated for decades. An overview about early works in
which ontologies are used to infer implicit information
about planning states can be found in~\cite{Gil2005}. Different approaches have
since then been used to
model planning domains, actions, and even planning problems using ontologies,
but also to use ontologies to generate planning problems, in domains as diverse as
kitting and assembly~\cite{BalakirskyKKPSG2013,KootballySLKG2015},
semantic web service decomposition~\cite{KluschGS2005,DurcikP2011},
robotics~\cite{AlsafiVMC007}, train depot management~\cite{LouadahPMTHB2022} and
manufacturing~\cite{BorgoCOU2019}.
These approaches usually depend on a static ontology
that is used to generate specifications for the planner, while the actions of
the planning specifications cannot modify the ontology. 

In~\cite{Milicic2007,ZarriessClassen2015}, actions can use DL concepts in the 
preconditions and postconditions of an action, which then operate on the models 
of an OWL ontology. A downside of letting actions directly operate on the models 
is that it is not trivial to determine the implicit consequences of an action,
that
is, to ensure that after executing an action on a model, we obtain an interpretation 
that is still a model of the ontology. This problem is also known as the ramification problem.
%
%

The ramification problem is avoided in approaches where actions do not operate on 
models, but on the knowledge base itself. 
This is the case with the 
\emph{Knowledge Action Bases} (KABs) and \emph{extended Knowledge Action
Bases} (eKABs) introduced in~\cite{HaririCMGMF2013,CalvaneseMPS2016}, which
combine DL knowledge bases
with actions that can add facts to and remove them from the knowledge
base. Here, every state in the planning domain corresponds to a DL knowledge base,
and pre-conditions of actions can query implicit information entailed in the
current state via DL reasoning. 
The idea 
is that
what is known about the world in each system state is represented using facts
of a knowledge base, interpreted as potentially incomplete under the
open-world assumption, and any implicit consequences of an action are accessed only through reasoning with the ontology.
Existing approaches to plan with eKABs practically rely on rewriting
the eKABs into planning problems in pure PDDL~\cite{GhallbHKMRV1998} or its 
extension
with derived predicates, i.e. axioms~\cite{HoffmannE2005},
so that a standard planning system such as Fast-Downward planner~\cite{Helmert2006}
can be
used. The limits of such an approach are investigated
in~\cite{BorgwardtHKKNS2022}, where the underlying ontology can be expressed in
the description logic \HornALCHOIQ, which roughly corresponds to the
Horn-fragment of OWL DL without complex object property axioms. While
\HornALCHOIQ is quite expressive, there are many properties useful for planning
that cannot be expressed (see Section~\ref{sec:framework} for a simple example).
To our
knowledge, no research in this direction considers more expressive ontology
languages.
%
%
%
%

Our approach is close to that of eKABs, but goes beyond existing approaches:
1) Rather than integrating actions and
knowledge, we strive for a separation of the representation
formalisms, and 2) using a domain-dependent rewriting approach, we are able to
support the full OWL~DL~2 syntax as defined in~\cite{W3COWL2012}
(including SWRL rules \cite{HorrocksPBTG2004}).

The aim of 1) is to have a presentation format that is tailored
towards the specific needs and skills of knowledge engineers and planning
experts. In particular, in our framework, we favor a strong separation of
concerns, with the planning specification encoded in standard PDDL, and the
domain knowledge encoded in a separate OWL ontology. The connection between
the two is established via an interface that links statements in the 
planning language to OWL axioms.
%
This way,
existing OWL ontologies can be easily integrated, and PDDL experts do
not need to learn another knowledge representation formalism.

Our solution to 2) is inspired by a technique for
ontology-mediated probabilistic model
checking presented in~\cite{OM_PMC,OM_PMC2}, which
uses a similar separation of concerns as our approach, but with a simpler
representation of states using propositional logic.
This allows us to support ontologies that go beyond Horn, and are thus able to use 
many naturally occurring constructs such as disjunction (e.g. to express that 
a valve must be either open or closed), or at-most constraints (e.g. to express how many 
objects an AUV can carry). 
Similar
to the work in~\cite{GocevGRPD2018}, we use justifications~\cite{Just} to
determine which elements of a planning state are relevant to an action
to be executed. However, while the authors of~\cite{GocevGRPD2018} are
interested in
explaining pre-conditions in an action for a singular state, we use
justifications to determine conditions on all possible states.

This paper extends our work on defining Ontology-Mediated planning as presented 
in \cite{OMPlanningPLATO} by a first evaluation. We demonstrate with the 
implementation that our method is capable of dealing with complex planning 
problems but that there are also planning domains where existing methods are 
superior.

\section{Preliminaries}

We recall the relevant notions regarding planning with PDDL. We assume the reader is familiar 
with the basics of OWL and description logics (DLs). For an introduction into OWL and description
logics, we refer to~\cite{DL_TEXTBOOK}. We further assume standard knowledge of first-order logic,
and use $\models$ to express entailment between theories and satisfaction in models. We call 
a formula $P(\vec{t})$ \emph{atom}, which is \emph{ground} if $\vec{t}$ contains only constants.

\subsection{PDDL Planning Specifications}
We consider the common syntax and semantics as introduced in
\cite{GhallbHKMRV1998,Lifschitz1987} and described in detail in 
\cite{FoxL2003}. A \emph{PDDL planning specification} $\PDDLcomponent$ is a 
tuple 
$\tup{\PDDLdomain,
\PDDLproblem}$ that contains a \emph{domain} $\PDDLdomain = \tup{\mathcal{P},
\mathcal{A}, \mathcal{D}}$ and a \emph{problem} $\PDDLproblem = \tup{O,
I, G}$. Here, 
$\mathcal{P}$ is a finite set of predicate names, $\mathcal{A}$ a finite set of
actions,
$\mathcal{D}$ a finite set of derivation rules,
${O}$ a finite set of objects, $I$ is an initial state and the goal $G$ is a 
first-order formula with predicates from $\mathcal{P}$. A \emph{state} is 
a finite set of ground atoms over $\mathcal{P}$ and
${O}$, interpreted as first-order interpretation; an \emph{action} is a tuple $a= \tup{V, \pre, \eff}$ where $V$ is a 
vector
of
variables, $\pre$ is the \emph{precondition} (a first-order formula with
predicates from $\mathcal{P}$ and free variables from $V$) and $\eff = 
\tup{\add, 
\del}$ is
the \emph{effect}. Both $\add$ and $\del$ are finite sets of atoms
over predicates from
$\mathcal{P}$ using variables from $V$ and constants from $O$. If neither 
$\pre$ nor $\eff$ contain
variables from $V$ or $V=\emptyset$, we call $a$ a \emph{ground action}.

\emph{Derivation rules} are of the form
$p(V)
\leftarrow \phi(V)$, where $V$ is a vector of variables, $p\in\mathcal{P}$,
and $\phi$ is a
first-order formula over the predicates
in $\mathcal{P}$ with free variables $V$ and constants from~$O$. We often call
derivation rules
just \emph{rules} and $\phi(V)$ the \emph{body} of a rule. If a predicate $p$
occurs on the left hand side of a rule, it is called a \emph{derived predicate}.
Derived predicates are neither allowed to occur
negatively
in a
derivation rule, nor are they allowed to occur in an effect of an action. For
a finite set of atoms $s$, we define $\mathcal{D}(s)$
as the least fix point over the possible applications of some rules from
$\mathcal{D}$ to the atoms in $s$, i.e., we apply the rules from $\mathcal{D}$
exhaustively and add the derived ground atoms until no more rules can be
applied.

Let $a=\tup{V, \pre, \eff}$ with $\eff = \tup{\add, \del}$ be an action and 
$\theta : V
\mapsto {O}$ a variable assignment. We denote by $\theta(a)$ the ground action
obtained by replacing each $x\in V$ in $a$ with $\theta(x)$. A ground action
is \emph{applicable} in a state $s$ iff
$\mathcal{D}(s) \models \pre$, that is, the precondition is evaluated over the 
atoms in the state and the entailed derived atoms. The result of applying the 
action $a$ on $s$ is then
denoted $s(a)$, defined as $s(a) := (s
\setminus \del) \cup \add$, i.e., all atoms are deleted and added according to
the effect. A \emph{plan} $\pi$ is now a sequence $a_1\ldots a_n$ of ground
actions that generates a sequence of states $s_0 \ldots s_{n}$ such that 1)
$s_0=I$ is the initial state of the planning problem, 2) for each
$i\in\{1,\ldots,n\}$, $a_i$ is applicable in $s_{i-1}$ and $s_i=s_{i-1}(a_i)$,
and 3) the goal is reached: $\mathcal{D}(s_n) \models G$.

There are many extensions to PDDL, for example conditional effects. The
described
components are the ones necessary for our framework but it can also be used
with such extensions.

\newcommand{\mycomment}[1]{}

\mycomment{
\subsection{OWL Ontologies}
For the context of this paper, the exact syntax and
semantics of OWL is not relevant. For an introduction into OWL and description
logics, we refer to~\cite{DL_TEXTBOOK}. The OWL standard defines different
\emph{profiles} that differ in expressiveness. We focus on the profile
\emph{OWL DL}, which is the most expressive profile that is still decidable. For
the purpose of this paper,
OWL DL can be seen as a fragment of first-order logic where only unary and binary
predicates can be used. An OWL ontology is then a collection of \emph{OWL
axioms},
which are formulated in this logic. In the context of OWL, unary predicates are
called \emph{OWL classes}, binary predicates are called \emph{OWL
properties}, and constants are called \emph{named individuals}. To ensure
unique naming across different ontologies, those are named using
\emph{Internationalized Resource Identifiers}, short \emph{IRIs}, which look
like URLs and are sometimes correspond to web address with further information.
\patrick{Handwavy and confusing - can we leave this detail out?}
There are many
syntaxes that can be used to represent OWL ontologies. \emph{Manchester syntax}
is a
syntax that is optimized for human writing on a computer keyboard, which is
what is used in the input files of our implementation. We also often
use the more mathematical and concise \emph{DL syntax}, which uses the syntax
of \emph{description logics}, the family of logics that forms
the formal foundation of OWL DL. OWL DL allows for a
large variety of class and axiom constructors. In DL syntax,
the most important axioms are of the forms
$C\sqsubseteq D$, $C\equiv D$, $a: C$ or $(a,b): r$, where $a$, $b$ are named
individuals, $r$ is an OWL property, and $C$, $D$ are \emph{class expressions},
which can be OWL classes or complex expressions describing sets of objects in
the way a first-order logic formula with one free variable would do it. Axioms
of the form $C\sqsubseteq D$ and $C\equiv D$ specify terminological knowledge
by putting constraints on the interpretations of OWL classes and OWL
properties. For
example, the axiom $A\sqcap\exists r.B\sqsubseteq C$ expresses that
if something is an instance of the OWL class $A$, and has a successor in the
$r$-relation that is an instance of $B$, then it must also be an instance of
$C$.  Axioms of the form $a: C$ and $(a,b): r$ specify assertional
knowledge about the specific individuals $a$ and $b$. Here, $a: C$
expresses that $a$ is an instance of $C$, and $(a,b): r$ corresponds to the
ground atom $r(a,b)$. In Manchester syntax, we write $a: C$ as 
\enquote{\texttt{a Type: C}}. 
If no complex class expressions are involved, assertional knowledge can be
nicely represented using graphs (see the bottom right of \Cref{fig:overview}
for an example).
Using an OWL reasoner, one can derive
implicit information in the form of OWL axioms from an OWL ontology. For an
ontology $\Omc$ and axiom $\alpha$, we write $\Omc\models\alpha$ if $\alpha$
can be derived from $\Omc$. We use $\Ind(\Omc)$ to
refer to the
named individuals that occur in $\Omc$.
}

\section{The Framework}\label{sec:framework}

We capture our framework formally via \emph{ontology-mediated planning
specifications}. At the heart of those is the notion of
\emph{ontology-enhanced states}, which
combine a PDDL state with an OWL ontology.

\begin{definition}[Ontology-Enhanced State]
 An \emph{ontology-enhanced state} is a tuple $q=\tup{P_q,\Omc_q}$, where $P_q$
is a set of atoms called the \emph{planner perspective of $q$}, and $\Omc_q$ is
a set of OWL axioms called the \emph{OWL perspective of $q$}.
\end{definition}

The
idea is that each state has a \emph{planner perspective}, on which the planner
directly operates, and on which preconditions and effects of actions are
evaluated and executed, respectively. The planner perspective of an 
ontology-enhanced state
is, as for classical planning problems, a set of ground atoms, where
predicates of arbitrary arity may occur. On the other side, there is the
\emph{OWL perspective} of the ontology-enhanced state, which corresponds to an
OWL
ontology, i.e. a set of OWL axioms, and from which implicit entailments can be
derived using reasoning.
The two perspectives are linked via an interface: which axioms are in the OWL
perspective depends on the atoms in the planner perspective. There is however
also a static part, which we call the static ontology, that describes
time-independent information (such as class definitions and general domain
knowledge), which is obtained
from an external OWL file and has no direct correspondence in the planner
perspective. The planner perspective can access implicit information from the OWL perspective 
using \emph{query predicates}.
%
Specifically,
whether a query-atom is active in the planner perspective depends on what can
be derived from the OWL perspective of the state. Before we give the formal
definition of how this works, we illustrate this idea with an example.

\begin{figure}
		\centering
	\resizebox{\textwidth}{!}{
	\begin{tikzpicture}[line width=0.75pt]
		\coordinate (origin) at (0,0);

		\draw (origin) node[individual, minimum width=2cm] (auv) {stackBot};


		\draw ($(auv) + (2.5,0)$) node[individual] (wp1) {blockA};
		\draw ($(wp1) + (1.4,0)$) node[individual] (wp2) {blockB};
		\draw ($(wp2) + (1.4,0)$) node[individual] (blockC) {blockC};

		\draw ($(auv) + (0,-1.5)$) node[class] (PR2) {\strut PR2};
		\draw ($(PR2) + (3.9, 0)$) node[class, minimum width=2.75cm] (WP) 
		{\strut 
		Block};

		\draw ($(PR2.south) + (2.25,-0.25)$) node[anchor=north] (TBox) 
		{\setlength{\tabcolsep}{2pt} \begin{tabular}{r c l}
			    $\owl{blockA} \not\simeq \owl{blockB}$& & $\owl{blockA} \not\simeq \owl{blockC}$\\
			    $\owl{blockB}$&$\not\simeq$&$\owl{blockC}$\\
				$\owl{PR2}$ &$\sqsubseteq$ &$\owl{Robot}\sqcap 
				{\leq}2\owl{holds}.\owl{Block}$ \\
				$\owl{PR2} \sqcap {=}2\owl{holds}.\owl{Block} $
				&$\sqsubseteq$ &$ \owl{FullHands}$ 
		\end{tabular}};

		\draw ($(auv) + (2.25, 2.75)$) node[textNode] (query)
{$\models \owl{FullHands} (\owl{stackBot}) $};

		\draw ($(query) + (-8,0)$) node[textNode] (PDDLquery) 
		{\pddl{fullHands(stackBot)}};

		\draw ($(auv) + (-5.75,1.75)$) node[textNode] (fluent) 
		{\pddl{holds(stackBot, blockB)}};
		\draw ($(fluent) + (0,-0.5)$) node[textNode] (fluentA) 
		{\pddl{holds(stackBot, blockA)}};

		\draw ($(fluentA) + (0,-1)$) node[textNode] (fluent2) 
		{\pddl{on(blockB, blockA)}};
		\draw ($(fluent2) + (0,-0.5)$) node[textNode] (fluent3) 
		{\pddl{onTable(blockC)}};

		\coordinate (onPath1) at ($(auv.north) + (2, 1.25)$);
		\coordinate (onPath2) at ($(onPath1) + (0.9, 0)$);

		\draw[->, rounded corners, relation, color=\Interfacecolor] 
		($(auv.north) + (0.75, 0)$) 
		-- 
		(onPath1) -- (onPath2) -- (wp2);
		
		\coordinate (onPath3) at ($(auv.north) + (0, 0.75)$);
		\coordinate (onPath4) at ($(onPath3) + (1.75, 0)$);
		
		\draw[draw=white,double distance=\pgflinewidth,ultra thick] 
		(onPath3) -- (onPath4);
		\draw[->, rounded corners, relation, color=\Interfacecolor] 
		($(auv.north) + (-0.75, 0)$) -- 
		(onPath3) -- (onPath4) -- (wp1);

		\coordinate (help6) at ($(auv) + (2.45, 0)$) ;
		\draw[color=black] (help6 |- fluent) node (relationLabel) {\owl{holds}};
		\coordinate (help14) at ($(auv) + (0.75, 0)$) ;
		\draw[color=black] (help14 |- fluentA) node (relationLabelA) 
		{\owl{holds}};

		\draw[->, relation] (auv) --node[left] {\footnotesize Type:} (PR2);


		\coordinate (blockAsoutheast) at (wp1.south east);
		\coordinate (blockAanker) at ($(blockAsoutheast) + (-0.2, 0)$);
		\draw[->, relation] (blockAanker) --node[left] 
		{\footnotesize 
		Type:} 
		(blockAanker |- WP.north);
		\draw[->, relation] (wp2.south) --node[left] {\footnotesize Type:} 
		(wp2.south |- WP.north);
		
		\coordinate (blockCsoutheast) at (blockC.south west);
		\coordinate (blockCanker) at ($(blockCsoutheast) + (0.2, 0)$);
		\draw[->, relation] (blockCanker) --node[left] 
		{\footnotesize 
			Type:} 
		(blockCanker |- WP.north);

		\draw[-latex, draw=\Interfacecolor] (fluent) --node[above] 
		(Fvertical) {\phantom{$F$}} ($(relationLabel.west) + (-0.25, 0)$);
		\draw[-latex, draw=\Interfacecolor] (fluentA) -- 
		($(relationLabelA.west) + (-0.25, 0)$);
		\draw[-latex, draw=\Interfacecolor] (query.west) -- node[above] 
		(Svertical) {\phantom{$S$}} (PDDLquery);


		\coordinate (help0) at (TBox.south east |- wp2.north east);
		\coordinate (help1) at ($(help0) + (1, 0.25)$);
		\draw[decoration={brace, raise=-5pt, amplitude=5pt}, decorate, draw=\OWLcolor]
		(help1) -- node[right=-5pt] {\footnotesize\begin{tabular}{c}
				static\\
				part of\\
				ontology
		\end{tabular}} (help1 |- TBox.south east);

		\coordinate (help2) at (help1 |- query.north);
		\coordinate (help3) at (help1 |- query.south);
		\draw[decoration={brace, raise=-5pt, amplitude=5pt}, decorate, draw=\OWLcolor]
		(help2) -- node[right=-5pt] {\footnotesize\begin{tabular}{c}
				ontology\\
				query
		\end{tabular}} (help3);

		\coordinate (help4) at ($(help3) + (0, -0.3)$);
		\coordinate (help5) at ($(help1) + (0, 0.1)$);
		\draw[decoration={brace, raise=-5pt, amplitude=5pt}, decorate, draw=\OWLcolor]
		(help4) -- node[right=-5pt] {\footnotesize \begin{tabular}{c}
				dynamic\\
				part of\\
				ontology
		\end{tabular}} (help5);

		\coordinate (help9) at (fluent.west |- PDDLquery.north);
		\coordinate (help10) at (fluent.west |- PDDLquery.south);
		\draw[decoration={brace, mirror, raise=5pt, amplitude=5pt}, decorate, 
		draw=\Planningcolor]
		(help9) -- node[left=5pt] {\footnotesize \begin{tabular}{c}
				query-\\
				atom
		\end{tabular}} (help10);

		\draw[decoration={brace, mirror, raise=5pt, amplitude=5pt}, decorate, 
		draw=\Planningcolor]
		(fluent.north west) -- node[left=5pt] {\footnotesize \begin{tabular}{c}
				mapped\\
				atoms
		\end{tabular}} (fluentA.south west);
	
		\draw[decoration={brace, mirror, raise=5pt, amplitude=5pt}, decorate, 
		draw=\Planningcolor]
		(fluent.west |- fluent2.north) -- node[left=5pt] {\footnotesize 
		\begin{tabular}{c}
				atoms\\
				outside\\
				mapping 
		\end{tabular}} (fluent.west |- fluent3.south);

		\coordinate (help11) at ($(query.north west) + (-1.5, 0)$);
		\draw[decoration={brace, raise=10pt, amplitude=5pt}, decorate, draw=\Planningcolor]
		(fluent.west |- query.north) -- node[above=15pt] 
		{\textbf{planning perspective}} (fluent.east |- query.north);
		\draw[decoration={brace, raise=10pt, amplitude=5pt}, decorate, draw=\OWLcolor]
		(help11) -- node[above=15pt] {\textbf{OWL perspective}} ($(help2) + (-0.5, 0)$);

		\coordinate (help7) at ($(fluent.east |- query.north) + (0.2, 0)$);
		\coordinate (help8) at ($(help11) + (-0.2, 0)$);
		\draw[decoration={brace, raise=10pt, amplitude=5pt}, decorate, draw=\Interfacecolor]
		(help7) -- node[above=17pt] (interface) {\textbf{interface}} (help8);
		
		\draw (interface |- Svertical) node {$S$};
		\draw (interface |- Fvertical) node {$F$};

		\coordinate (help12) at (help11 |- help1);
		\draw[dotted, draw=\OWLcolor, line width=1.25pt] ($(help1) + (-0.5, 0.05)$) -- ($(help12) + (-0.0, 0.05)$);

		\coordinate (help13) at (help11 |- help4);

		\draw[draw=\OWLcolor,double] ($(help4) + (-0.5, 0.15)$) -- ($(help13) + (-0.0, 0.15)$);

	\end{tikzpicture}
}
	\caption{Example of ontology based planning. The interface maps ontology
		queries to planning predicates and atoms in the planning perspective to 
		ABox atoms. 
		The static part of the ontology contains information about instances 
		(ABox) as
		well as general axioms (TBox). The connections between
the two perspectives via
		the fluent (F) and query (S) interface are shown in green.}
	\label{fig:overview}
\end{figure}
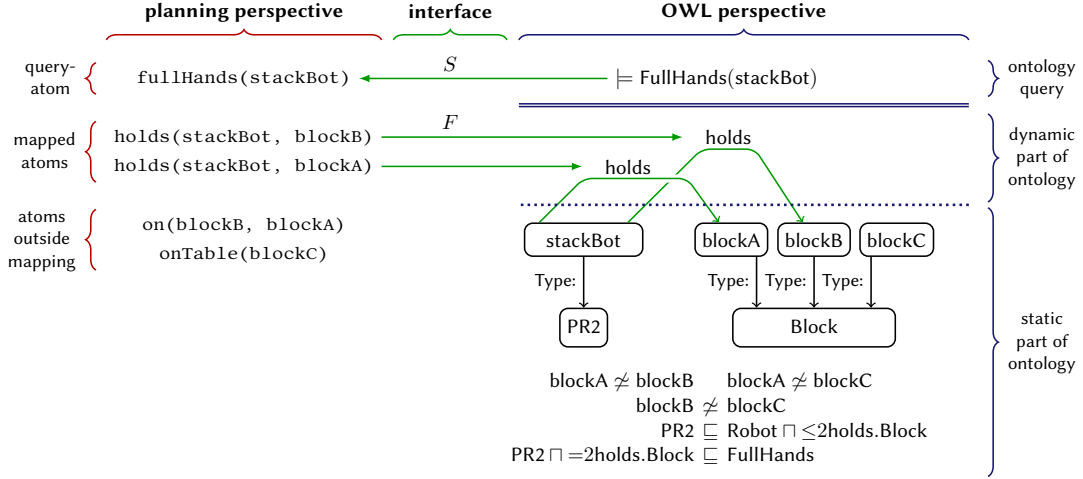

\begin{example}\label{example:overview}
An example of an ontology-enhanced state is depicted in
Figure~\ref{fig:overview}. 
The scenario is inspired from the classical blocksworld planning example. In
contrast to the classical problem where the robot has only one hand, we use an
OWL ontology to specify the type of the robot and infer its number of hands. In 
the 
example, the stacking robot is a \owl{PR2} robot \cite{BohrenRGMP2011} that can hold 
two blocks at a time, and if it holds two blocks, it becomes an instance of 
\owl{FullHands}. While relatively simple, those cardinality constraints
already go beyond the expressivity of \HornALCHOIQ, the most expressive DL
currently supported by existing
implementations for eKABs (see Section~\ref{sec:related-work}).
%
The planner perspective of the state is shown on the left, and
the OWL
perspective is shown on the right. The interface is in the
middle. If the atom
\pddl{holds(stackBot, blockA)} becomes true in the planner perspective, 
this
is reflected in the ontology perspective as an OWL axiom
expressing a corresponding relation between the two
individuals \owl{stackBot} and \owl{blockA}. Using the static 
ontology, we can infer that \owl{stackBot} is an instance of the OWL
class \owl{FullHands}, because the \owl{holds} relation is 
true 
for two different blocks.
This is reflected by the entailed OWL axiom
\owl{FullHands(stackBot)}. We also have a query predicate
\pddl{fullHands}, which corresponds to a query over instances
of the OWL class \owl{FullHands}. Since we can infer from the OWL
perspective that \owl{stackBot} is an instance of 
\owl{FullHands},
the atom \pddl{fullHands(stackBot)} becomes true in the
planner perspective of the state.
\end{example}

The central notion of this paper is that of an \emph{ontology-mediated planning specification},
which consists of the following three components:
\begin{enumerate}
	\item the \emph{PDDL component} $\PDDLcomponent$, which is a PDDL
planning specification consisting of a domain and a problem,
	\item the \emph{static ontology} $\Omc$, which is an OWL ontology
specifying the static knowledge, that is, it contains axioms whose truth
cannot
be affected by actions, and
	\item the \emph{interface} that specifies how the two
perspectives of an ontology-enhanced state should be linked. The interface
itself consists of two parts:
	\begin{enumerate}
		\item the fluent interface, and
		\item the query interface.
	\end{enumerate}
\end{enumerate}

\begin{figure}
	\setlength\fboxsep{10pt}
	\mbox{} \hfill
	\begin{subfigure}[b]{0.49\textwidth}
		\centering
		\begin{boxedverbatim}
 OBJECT    stackBot   -> stackBot
 OBJECT    blockA     -> blockA
 OBJECT    blockB     -> blockB
 OBJECT    blockC     -> blockC

 PREDICATE holds(_,_) -> holds
		\end{boxedverbatim}
		\caption{Example of a fluent interface.}
		\label{fig:fluent-specification}
	\end{subfigure}
	\hfill
	\begin{subfigure}[b]{0.49\textwidth}
		\centering
		\begin{boxedverbatim}
 PREDICATE: fullHands
 VARIABLES: ?r
 TYPE_SPECIFICATION:
    Robot(?r)
 QUERY:
    FullHands(?r)
		\end{boxedverbatim}
		\caption{Example of a query specification.}
		\label{fig:query-specification}
	\end{subfigure}
	\hfill \mbox{}
	\caption{Interface specification using the syntax of our implementation.}
\end{figure}

The
\emph{fluent interface} maps objects, unary and binary predicates used in the
planner perspective to
the named individuals, OWL classes and OWL properties
that are used in the OWL perspective.
An example of how this looks like for our
implementation is shown in Figure~\ref{fig:fluent-specification}.
In the context of this paper, it is convenient to see the fluent
specification simply as a partial function $F$ that assigns to some of the
predicates and objects $X$ in
the planning specification an IRI $F(X)$. We require $F$ to be inverse
functional,
that is, $F^-$ is also a function. We lift $F$ in a straight-forward way to
atoms by setting $F(P(t_1,\ldots,t_n))=F(P)\big(F(t_1),\ldots,F(t_n)\big)$ if it
is defined.

The \emph{query interface} is a set of \emph{query specifications}
$S=\tup{p_S, V_S, T_S, Q_S}$, which each consist of four components:
\begin{enumerate}
 \item $p_S$ is the \emph{query predicate},
 \item $V_S$ is a vector of \emph{query variables}, whose number corresponds
to the arity of $p_S$,
 \item the \emph{type specification} $T_S$ assigns to each variable $x\in V_S$
an OWL class expression
specifying its \emph{static type}, and
 \item the \emph{query} $Q_S$ is a set of OWL axioms using variables from $V_S$
as place holders
for individual names.
\end{enumerate}

An example of how this looks like for our implementation is shown in
\Cref{fig:query-specification}.
Note that in the type specification, we can only assign
one class expression to each variable, while variables may occur in arbitrary
ways in the query. The static types are used to restrict the
set of named individuals that can be assigned to a variable: candidates for a
variable
$x\in V_S$ are individual names $a$ for which the static ontology entails
$a: T_S(x)$, that is, which are an instance of the class expression assigned to
$x$ via $T_S$. For the specification in \Cref{fig:query-specification},
$V_s=(\textnormal{?r})$ and ?r can be associated with instances of 
the
class \owl{Robot}.
For a given static ontology $\Omc$ and query specification $S$, we thus have
a set
$\Theta(S,\Omc)$ of \emph{legal assignments} $\theta:V_S\rightarrow\Ind(\Omc)$ 
of
variables to individual names in $\Omc$. Finally, $Q_S$ specifies the OWL query that 
the query predicate $p_S$ stands for. For a given assignment $\theta\in
\Theta(S,\Omc)$, $\theta(Q_S)$ denotes the set of OWL axioms obtained by 
replacing each
variable $x\in V_S$ in $Q_S$ by $\theta(x)$. In the present example, for the
assignment $\theta(\textnormal{?r})= \owl{stackBot}$,
we would have $\theta(Q_S)=\{~ \owl{fullHands(stackBot)}~ \}$.

\newcommand{\Sbf}{\mathbf{S}}

We have now all ingredients to define ontology-mediated planning specifications.

\begin{definition}
 An \emph{ontology-mediated planning specification} is a tuple
$\tup{\PDDLcomponent,\Omc,F,\Sbf}$, where $\PDDLcomponent$ is a PDDL planning
specification consisting of a planning domain and a planning problem, $\Omc$ is
an OWL ontology called the \emph{static ontology}, $F$ is a
fluent interface, and $\Sbf$ is a set of query specifications called the
\emph{query interface}.
\end{definition}

An ontology-mediated planning specification determines when an
ontology-enhanced state is compatible for that specification. In particular, a 
state 
$q=\tup{P_q,\Omc_q}$ is \emph{compatible} to an ontology-mediated planning
specification $\textbf{OP}=\tup{\PDDLcomponent,\Omc,F,\Sbf}$, where
$\Dmc$ are the derivation rules in $\PDDLcomponent$, iff:
\begin{enumerate}[label=\textbf{C\arabic*},leftmargin=*]
 \item\label{comp:state} $P_q$ is a set of atoms over predicates and constants
occurring in $\PDDLcomponent$,
 \item\label{comp:static} $\Omc\subseteq\Omc_q$ (the static ontology is always
part of the
OWL perspective),
 \item\label{comp:fluents} for every atom $\alpha\in \Dmc(P_q)$ for which
$F(\alpha)$ is defined,
$F(\alpha)\in\Omc_q$
 \item\label{comp:minimal} $\Omc_q$ contains no axioms that are not required
due to Conditions~\ref{comp:static} and~\ref{comp:fluents}
 \item\label{comp:query} for every query specification
$S=\tup{p_S,\tup{x_1,\ldots,x_n},T_S,Q_S}\in\Sbf$ and
 $\theta\in
\Theta(S,\Omc)$, if $F^-(\theta(x_i))$ is defined for each variable $x_i$ and
$\Omc_q \models \theta(Q_S)$, then 
$$p_S(F^-(\theta(x_1)),\ldots,F^-(\theta(x_n)))\in 
P_q.$$
\end{enumerate}

\newcommand{\ext}{\textsf{ext}}

Given an ontology-mediated planning specification
$\textbf{OP}=\tup{\PDDLcomponent,\Omc,F,\Sbf}$ and a state $P$ in the
corresponding planning domain, we define the
\emph{extension $\ext(P,\textbf{OP})$ of $P$ according to $\textbf{OP}$}
as follows. 
Let 1) $P'$ be the set of
atoms in $P$ that are not over query predicates, 2) $\Omc_q$ the set of axioms
required to satisfy Conditions~\ref{comp:static} and~\ref{comp:fluents} based
on the
atoms in $P'$, and 3) $P_q$ the extension of $P'$ by all atoms over query
predicates that are required to satisfy Condition~\ref{comp:query} for the
ontology $\Omc_q$. Then, $\ext(P, \textbf{OP})=\tup{P_q,\Omc_q}$.


\begin{example}
	Consider the example in \Cref{fig:overview} where $\alpha =
\pddl{holds(stackBot, blockA)}$ and $\beta =
\pddl{holds(stackBot, blockB)}$ with $\alpha, 
\beta \in 
P_q$ and $F$ is 
defined as in
\Cref{fig:fluent-specification} and \textnormal{S} as in
\Cref{fig:query-specification}. Then, according to \ref{comp:fluents}, the
axioms from the mappings
$F(\alpha) = \owl{holds(stackBot, 
blockA)}$ 
and $F(\beta) = 
\owl{holds(stackBot, blockB)}$ are part
of $\Omc_q$. 
Using the static part of
$\Omc_q$, which states that
\owl{stackBot} is a PR2 robot and  \owl{blockA} is 
different from \owl{blockB}, we can
infer that $\Omc_q \models \{\owl{FullHands(stackBot)} 
\}$. 
Using 
$\theta =
\{(\textnormal{?r} \mapsto \owl{stackBot})\}$, $F$ and $S$ from
Figure~\ref{fig:query-specification}, we can apply \ref{comp:query} to determine
that
$\pddl{fullHands(stackBot)}\in P_q$.
\end{example}

It remains to define the semantics of actions and plans on ontology-mediated
planning specifications. Fix an ontology-mediated planning specification
$\textbf{OP}=\tup{\PDDLcomponent,\Omc,F,\Sbf}$. Let $a$ be a ground action
with precondition $\pre$ and effect $\eff=\tup{\add,\del}$. Let $q$ be an
ontology-enhanced state. We say that $a$ is \emph{applicable} on $q$ iff
$\Dmc(P_q)\models\pre$. The result of \emph{applying $a$ on $q$} is then denoted
by $q(a)$ and defined as $q(a)=\ext(P_q(a),\textbf{OP})$. We can now define
\emph{plans for \textbf{OP}} similarly as we did for planning specifications:
Namely, a plan is a sequence $a_1\ldots a_n$ of actions that generates a
sequence
$q_0q_1\ldots q_n$ of ontology-enhanced states s.t.
\begin{enumerate}
 \item $q_0=\ext(I,\textbf{OP})$, where $I$ is the initial state of the
PDDL planning problem in $\textbf{OP}$,
 \item for each $i\in\{1,\ldots,n\}$, $q_{i}=q_{i-1}(a_i)$,
 \item for each $i\in\{1,\ldots,n\}$, $a_i$ is applicable on $q_{i-1}$, and
 \item $\Dmc(P_{q_n})\models G$, where $\Dmc$ are the derivation rules of 
 the
planning domain, and $G$ is the formula describing the goal of the planning 
problem.\footnote{Note that we allow the plan to go through states whose OWL
perspective is inconsistent. If this is not wanted, an easy way to avoid this
would for example be to use a query predicate to detect such states, and to
adapt the preconditions of all actions so that they are not applicable in
inconsistent states. As a consequence, a goal state can never be reached from such a state.}
\end{enumerate}

\section{Solving Ontology-Mediated Planning Problems in Practice}

Semantically, our approach is very related to that of eKABs introduced
in~\cite{CalvaneseMPS2016}. eKABs do not offer a differentiation between OWL
perspective and
planner perspective. Instead, actions operate directly on OWL axioms, which can
be directly referenced to both pre-conditions and post-conditions of the
actions. We conjecture that it is always possible using simple transformations
to translate an eKAB with a finite domain into an ontology-mediated planning
problem. In the other
direction, we can translate ontology-mediated planning problems into eKABs by
replacing atom predicates by the corresponding OWL class and OWL properties,
and replacing query atoms by the corresponding queries. It is thus in theory
possible to use an eKAB planner to compute plans for ontology-mediated planning
problems. However, existing implementations for eKAB planning have limitations
regarding the supported OWL fragment. The general idea of these approaches is
to take the eKAB planning specification, and translate it into a PDDL
specification that can then be used by a standard PDDL planner. Those
techniques focus on the \emph{planning domain}, that is, the obtained
rewritings are independent of the planning problem. The approach
presented in~\cite{CalvaneseMPS2016,BorgwardtHKS2021} only supports
\emph{rewritable DLs}, which would
correspond to the OWL fragment OWL-QL. The approach presented
in~\cite{BorgwardtHKKNS2022} goes
further by using derivation rules, which allows to encode
\HornALCHOIQ via a known translations of such ontologies into datalog programs.
\HornALCHOIQ roughly corresponds to the Horn fragment of OWL DL. For DLs that
are not Horn, a translation into datalog is generally not possible, since
datalog is itself a Horn logic. The same applies to rewriting into derivation
rules, if those are supposed to be defined independently of the objects of the
planning problem.
Therefore, in order to support full OWL DL, we need to
take into account also the planning problem. Specifically, our approach directly
iterates over the possible
assignments for each query predicate. This allows us to develop a more
generic approach that does not restrict the ontology language, as long as a
reasoner for it is available.

The basic idea is to construct a derivation rule for each query predicate, which
determines for each valid variable assignment a set of
conditions that can be evaluated directly on the planner
perspective of a state.
The details on how we construct these derivation rules in practice can be found in the appendix.
\begin{figure}
	
	\footnotesize

	\begin{framed}
		\vspace{-0.4cm}
		\begin{align*}
			\texttt{inconsistent} \leftarrow& \left(\begin{array}{l l}
				&\pddl{holds(stackBot, blockA)}\\
				\land &\pddl{holds(stackBot, blockB)}\\
				\land& \pddl{holds(stackBot, blockC)}\\
			\end{array} \right)\\ \\
			\texttt{fullHands(stackBot)} \leftarrow& \pddl{inconsistent} \lor\\
			& \left( \begin{array}{l l}
				& (\pddl{holds(stackBot, blockA)} \land \pddl{holds(stackBot, 
				blockB)}) \\
				\lor & (\pddl{holds(stackBot, blockA)} \land 
				\pddl{holds(stackBot, blockC)}) \\
				\lor & (\pddl{holds(stackBot, blockB)} \land 
				\pddl{holds(stackBot, blockC)}) \\
			\end{array}\right)
		\end{align*}
		\vspace{-0.4cm}
	\end{framed}

	\caption{Computed Derivation rules resulting for the example from 
	Figure~\ref{fig:overview}.}
	\label{fig:derived-predicates}
\end{figure}

\begin{example}
	\Cref{fig:derived-predicates} depicts the generated derived predicates
for our running example. We introduce the atom \texttt{inconsistent}, which 
captures the states in the ontology perspective that are inconsistent. The atom 
is used in the derivation rule for every query atom. 
In our example, 
the static ontology states that every 
individual from the class \owl{PR2} is only allowed to hold at most 
two blocks.
Using the fluent interface, we can determine the combination of atoms in
the planning perspective that would lead to an ontology that would violate this 
constraint.
There is only one derivation rule for the query-predicate \pddl{fullHands} as 
the only possible variable mapping is $\textnormal{?r} = \pddl{stackBot}$ 
because \owl{stackBot} is the only individual with the static type \owl{Robot}. 
The query atom is true if the OWL perspective is inconsistent or if the 
\owl{stackBot} holds exactly two different blocks.

\end{example}

\section{Evaluation}
\paragraph{Implementation.}
We implemented our method of compiling ontology-mediated planning
specifications into PDDL specification with derivation rules.\footnote{The
source files and scripts to reproduce the evaluation can be obtained online \cite{SupplementaryMaterialForDL2023}.} 
We use the standard formats PDDL and TTL for the planning specification and
the ontology respectively, and we use our own text-based formats for the fluent 
and query interface. Our compilation algorithm relies on an extensive computation of
justifications, for which we used a modified version of the blackbox justification
algorithm implemented in the OWL-API~\cite{OWL_API}, together
with the OWL reasoning system
HermiT~\cite{GlimmHMSW2014}. The computed derivation rules are added to the PDDL
domain. We used the fast-downward planning system \cite{Helmert2006} with the
heuristic A* for planning. We chose this heuristic because many of the more advanced heuristics have problems working with derivation rules.

For our evaluation, we compare our method to the eKAB method presented in 
\cite{BorgwardtHKKNS2022}. We choose this competitor as it is the 
implementation that can deal with the most expressive DL fragment and performs 
best on existing benchmark domains \cite{BorgwardtHKKNS2022}. We used a time 
limit of 1200s and a memory limit of 8GB. Both limits applied to compiling and 
planning individually. 

\paragraph{Benchmarks.}
Our benchmark consists of instances from the domains used in~\cite{BorgwardtHKKNS2022}, as well as some new domains. As to be expected, our method is at this stage not yet competetive on all domains, and in fact, on some
of the domains used in~\cite{BorgwardtHKKNS2022} to evaluate the performance of 
eKABs based on rewritable and Horn DLs, our method almost always timed out.
To
have a more interesting picture, we focus here on the more complex domains from 
that paper (``Drones'' and ``Queens''), which surprisingly turned out also to 
be the more interesting ones for our approach, and present the other results in 
the appendix.
In particular, our benchmark set contained 54 instances from two of the most complex domains that were introduced in \cite{BorgwardtHKKNS2022},
%
to which we added two new domains with
39 instances. The existing instances are eKABs, which are based on the DL fragment
\HornALCHOIQ. We translated them manually to ontology-mediated planning
specifications, which mainly involved specifying the interface. The domains
\enquote{Pipes} and \enquote{Blocksworld} were created by us. \enquote{Pipes} is
a complex domain describing a mission for an underwater robot in a 2D world. The
world contains pipes, valves and tanks that can be connected to each other and
that are located at different waypoints. The goal is to document damages of the
pipe and to turn the valves such that no tank is connected to a damaged pipe
segment. \enquote{Blocksworld} reflects the domain from our running example (see
e.g. Example~\ref{example:overview}). It is inspired by the Blocksworld domain
from the international planning competition 2000 \cite{Bacchus2001}. This domain
uses axioms that can not be captured by \HornALCHOIQ.

\paragraph{Results.}

\begin{table*}
	\caption{Results grouped by domain. Computation times are in seconds and the median over the commonly solved instances (except times marked with \enquote{*}, those refer to the median for the instances solved by this method). Best results are marked in bold (where applicable).} 
	\label{tab:eval}
	\begin{adjustbox}{max width=\textwidth}
	\begin{tabular}{lrr|rr|rr|rr|rr}
		\toprule
		& & \# axioms & \multicolumn{2}{c|}{\# solved} & \multicolumn{2}{c|}{\# compiled} & \multicolumn{2}{c|}{planning time} & \multicolumn{2}{c}{compilation time} \\
		Domain & \# & in T-Box & \HornImpl & \InterfaceImpl & \HornImpl & \InterfaceImpl & \HornImpl & \InterfaceImpl & \HornImpl & \InterfaceImpl \\
		\midrule
		
		Drones & 24 & 17 & \textbf{24} & 4 & \textbf{24} & 4 & 8.2\phantom{*} & \textbf{4.1}\phantom{*} & \textbf{0.6}\phantom{*} & 917.5\phantom{*}\\
		
		Queens & 30 & 17 & \textbf{24} & 19 & \textbf{30} & 20 & 3.0\phantom{*} & \textbf{0.8}\phantom{*} & \textbf{0.7}\phantom{*} & 146.5\phantom{*}\\

		Pipes & 21 & 43 & 14 & \textbf{19} & \textbf{21} & \textbf{21} & 5.4\phantom{*} & \textbf{0.2}\phantom{*} & \textbf{0.7}\phantom{*} & 10.0\phantom{*}\\
		
		Blocksworld & 21 & 5 & --- & 13 & --- & 21 & ---\phantom{*} & 4.8* & ---\phantom{*} & 1.1* \\

		\bottomrule
	\end{tabular}
	\end{adjustbox}
\end{table*}

Table~\ref{tab:eval} provides a summary of our experiments. We call the method
presented in this paper \InterfaceImpl and the method presented in
\cite{BorgwardtHKKNS2022} \HornImpl. \InterfaceImpl was capable of handling some
of the domains very well, while the performance on others is worse than
\HornImpl.

In general, \InterfaceImpl had longer compilation times and shorter planning times compared to \HornImpl. The derivation rules generated by \InterfaceImpl{} had a much simpler structure than the ones generated by \HornImpl because they only used atoms and no other derived predicates in the body. This resulted in a faster search in the planning phase as each state could be evaluated faster, e.g. in the domain \enquote{Pipes} the planner could, on average, evaluate 11,000 states per second for \HornImpl and 117,000 states per second for \InterfaceImpl. Therefore, we expected \InterfaceImpl to outperform \HornImpl in cases where the planner needs to search in a huge state space, while the number of fluents and queries is low. This is e.g. the case for the larger instances from the domain \enquote{Pipes}, which could be solved by \InterfaceImpl but not by \HornImpl.

The size of the ontology is in general not a problematic factor for \InterfaceImpl as the domain \enquote{Pipes}, which contains a larger T-Box than the other domains, could be compiled in rather short time. Similarly, increasing the expressiveness of the underlying DL does not seem have a negative effect, as all instances from the domain \enquote{Blocksworld} could be compiled within the provided bounds.

On the other hand, as the domain \enquote{Drones} shows, the performance on instances from the existing domains is often poor. As mentioned before, this picture was even worse with the other benchmarks from~\cite{BorgwardtHKKNS2022}, on which our method almost always caused a timeout.
One reason is that we need to map every atom from the planning perspective to the OWL perspective to describe equivalent instances to the eKAB instances. This results in many, often several hundred, fluents which again results in many explanations for an inconsistent ontology. As \InterfaceImpl enumerates all the possibilities in the derivation rules, this is a problem and leads to a huge increase in compilation time. The detailed evaluation in the appendix shows that this can happen even in relatively small instances.

\section{Conclusion}

We proposed ontology-mediated planning specifications as a way to integrate
OWL reasoning into planning. One objective was to find a formalism that allows
for a separation of concerns, allowing to separate the specification of 
ontologies from the
specification of planning problems and domains. This has the advantage that the
ontology can be maintained by ontology experts, while the planning specification
can be developed by planning experts, with the interface serving as the only
connecting component. We developed a first practical method for computing plans
for such planning problems, which relies on justifications. This technique
allows us to be flexible with respect to the ontology language, with the result
that our method supports the entirety of OWL DL, going beyond what is currently
supported by implementations for the related frameworks of KABs and eKABs.
Our evaluation shows that our method can outperform existing methods on some instances but is not competitive for most existing benchmark domains yet.
In the future, we want to investigate optimizations of our
approach, maybe combining it ideas of the other rewriting-based approaches,
in order to obtain shorter compilation times.

\begin{acknowledgments}
	Tobias John is part of the project REMARO that has received funding from the European Union's Horizon 2020 research and innovation programme under the Marie Sk\l{}odowska-Curie grant agreement No 956200. Patrick Koopmann is supported by the German Research Foundation (DFG),
	grant 389792660 as part of TRR 248 – CPEC. 
\end{acknowledgments}

\bibliography{bib-dl-planning}

\newpage

\appendix

\section{Solving Ontology-Mediated Planning Problems in Practice}

We solve ontology-mediated planning problems by generating a deriviation rule
for each query predicate in the interface.
To construct these derivation rules automatically, we make use of
\emph{justifications}, which were originally developed as a means
to explain entailments in ontologies to end-users~\cite{Just}. More specifically,
we use justifications relative to a second set of axioms, as defined in the following.

\newcommand{\Jmc}{\ensuremath{\mathcal{J}}\xspace}
\newcommand{\Just}{\text{Just}}

\begin{definition}
 Let $\Omc$, $\Omc'$ be ontologies s.t. $\Omc'\subseteq\Omc$ and $\alpha$ be an
axiom s.t. $\Omc\models\alpha$. Then, a \emph{justification for
$\Omc\models\alpha$ relative to $\Omc'$} is a set $\Jmc\subseteq\Omc$ s.t. $
\Jmc\cup\Omc'\models\alpha$ and for no $\Jmc'\subset\Jmc$,
$\Jmc'\cup\Omc'\models\alpha$. If $\Omc'=\emptyset$, $\Jmc$ is called a
\emph{classical justification}.
\end{definition}

We use $\Just(\Omc,\Omc',\alpha)$ to denote the set of all justifications of
$\Omc\models\alpha$ relative to $\Omc'$. A special case of justifications are
those of $\Just(\Omc,\Omc',\top\sqsubseteq\bot)$. This set
contains all
sets of axioms from $\Omc$ that are inconsistent with $\Omc'$.

Classical justifications can be
computed using standard tools such as the OWL API~\cite{OWL_API}.
To compute relative justifications, we use an adaption of the existing
implementation from the OWL API that we already used in~\cite{OM_PMC}.
This can now be used to compute the
described derivation rules. Fix an ontology-mediated planning specification
$\tup{\PDDLcomponent,\Omc,F,\Sbf}$. Let $\textbf{F}$ be the set of all OWL
axioms in the range of $F$, which we call \emph{fluents} from now on.
Furthermore, set
$\Just_\bot=\Just(\textbf{F}\cup\Omc,\Omc,\bot)$, and for an axiom
$\alpha$, we set
$\Just_\alpha=\Just(\textbf{F}\cup\Omc,\Omc,\alpha)\setminus\Just_\bot$.
$\Just_\bot$ contains all sets of fluents that are inconsistent with the static
ontology. $\Just_\alpha$ contains all sets of fluents that are consistent with
the static ontology, and lead to an entailment of $\alpha$ when added to it.
Since an inconsistent ontology entails every axiom, $\Just_\bot$ describes the
special case in which all query predicates should become active for all
assignments.

First, we construct a derivation rule for an atom
\texttt{inconsistent}:
\[
\pddl{inconsistent} \leftarrow
\bigvee_{\Jmc\in\Just_\bot}\bigwedge_{\alpha\in\Jmc}
F^-(\alpha) .
\]
Next, for a given query specification
$S=\tup{p_S,\tup{x_1,\ldots,x_n},T_S,Q_S}$,
and an assignment $\theta \in \Theta(S,\Omc)$, we define the formula
$\phi_{S,\theta}$ that
describes when $p_S$ should become active under that assignment. Specifically,
1) we check that all variables are assigned according to $\theta$ and the fluent
interface $F$, 2) we iterate over all the axioms $\alpha\in \theta(Q_S)$, 3) we
iterate over all the justifications $\Jmc\in\Just_\alpha$, and 4) we translate
all the axioms in $\Jmc$ based on the fluent interface $F$. This leads to the
following formula:
\[
 \phi_{S,\theta}=\bigwedge_{i=1}^n \left(x_i=F^-(\theta(x_i))\right)
 \wedge
 \bigwedge_{\alpha\in \theta(Q_S)}
\bigvee_{\Jmc\in\Just_\alpha}
\bigwedge_ {\beta\in \Jmc }F^-(\beta) .
\]
The derivation rule for $p_S(x_1,\ldots,x_n)$ is then constructed as follows:
\[
 p_S(x_1,\ldots,x_n) \leftarrow \pddl{inconsistent}\vee\bigvee_{\theta\in
 \Theta(S,\Omc)}\phi_{S,\theta}
\]

We use a special atom for inconsistent states to
simplify the derive-directives. But having such an atom has the further
advantage that we can easily adapt the planning specification to avoid
inconsistent states all together: we can for instance add a precondition to
every action that the current state is not inconsistent. This behavior is in
line with the existing semantics of eKABs \cite{CalvaneseMPS2016}.

The ontology-mediated planning specification is now translated to a standard
PDDL specification by just adding to \PDDLcomponent all these derivation rules.
We can now use an off-the-shelf planner to determine a plan for it.

\section{Evaluation Results on the Complete Benchmark}

\begin{table*}
	\caption{Results grouped by domain. Computation times are in seconds and 
	the median over the commonly solved instances (except times marked with 
	\enquote{*}, those refer to the median for the instances solved by this 
	method). Best results are marked in bold (where applicable).} 
	\label{tab:evalDetails}
	\begin{adjustbox}{max width=\textwidth}
		\begin{tabular}{lrr|rr|rr|rr|rr}
			\toprule
			& & \# axioms & \multicolumn{2}{c|}{\# solved} & 
			\multicolumn{2}{c|}{\# compiled} & \multicolumn{2}{c|}{planning 
			time} & \multicolumn{2}{c}{compilation time} \\
			Domain & \# & in T-Box & \HornImpl & \InterfaceImpl & \HornImpl & 
			\InterfaceImpl & \HornImpl & \InterfaceImpl & \HornImpl & 
			\InterfaceImpl \\
			\midrule
			Drones & 24 & 17 & \textbf{24} & 4 & \textbf{24} & 4 &
			8.2\phantom{*} & \textbf{4.1}\phantom{*} & \textbf{0.6}\phantom{*} 
			& 917.5\phantom{*}\\
			
			Queens & 30 & 17 & \textbf{24} & 19 & \textbf{30} & 20 & 
			3.0\phantom{*} & \textbf{0.8}\phantom{*} & \textbf{0.7}\phantom{*} 
			& 146.5\phantom{*}\\
			
			RobotConj & 20 & 22--174& \textbf{20} & 1 & \textbf{20} & 1 & 0.3* 
			& 0.12* & 0.8* & 114.6*\\
			Pipes & 21 & 43 & 14 & \textbf{19} & \textbf{21} & \textbf{21} & 
			5.4\phantom{*} & \textbf{0.2}\phantom{*} & \textbf{0.7}\phantom{*} 
			& 10.0\phantom{*}\\
			
			Blocksworld & 21 & 5 & --- & 13 & --- & 21 & ---\phantom{*} & 4.8* 
			& ---\phantom{*} & 1.1* \\
			Cats & 20 & 8 & \textbf{20} & 0 & \textbf{20} & 0 & 0.3* &
			---\phantom{*} & 0.8* & ---\phantom{*}\\

			Elevator & 20 & 9 & \textbf{20} & 1 & \textbf{20} & 1 & 2.3* & 0.3*
			& 0.8* & 264.7*\\

			Robot & 20 & 22--174 & \textbf{20} & 1 & \textbf{20} & 1 & 0.9*  &
			0.1* & 0.9* & 106.8*\\

			TaskAssign & 20 & 17 & \textbf{20} & 0 & \textbf{20} & 0 & 1.8* &
			---\phantom{*} & 0.8* & ---\phantom{*}\\

			TPSA & 15 & 29 & \textbf{5} & 0 & \textbf{15} & 0 & 108.8* &
			---\phantom{*} & 0.7* & ---\phantom{*} \\

			VTA & 15 & 6 & \textbf{15} & 0 & \textbf{15} & 0 & 13.1* &
			---\phantom{*} & 0.8* & ---\phantom{*} \\

			VTA-Roles & 15 & 14 & \textbf{15} & 0 & \textbf{15} & 0 & 15.2* &
			---\phantom{*} & 1.4* & ---\phantom{*} \\

			\bottomrule
		\end{tabular}
	\end{adjustbox}
\end{table*}

Table~\ref{tab:evalDetails} contains the result for the extended benchmark that
contains all domains from~\cite{BorgwardtHKKNS2022}, including those on which
our method almost always created a timeout.
Our observation is that in these domains, too many justifications for inconsistencies
could be found, so that our method was not able to compute the derivation rules 
within
the time limit.

\end{document}